\documentclass[sigconf, screen=true, review=False,anonymous=False,balance=False,nonacm]{acmart}
\renewcommand\footnotetextcopyrightpermission[1]{} % removes footnote with conference information in first column
\AtBeginDocument{%
	\providecommand\BibTeX{{%
			\normalfont B\kern-0.5em{\scshape i\kern-0.25em b}\kern-0.8em\TeX}}}

\usepackage{subfigure}
\usepackage{stfloats}
\begin{document}

%%
%% The "title" command has an optional parameter,
%% allowing the author to define a "short title" to be used in page headers.
\title{Translate the Facial Regions You Like Using Region-Wise Normalization}

%%
%% The "author" command and its associated commands are used to define
%% the authors and their affiliations.
%% Of note is the shared affiliation of the first two authors, and the
%% "authornote" and "authornotemark" commands
%% used to denote shared contribution to the research.

%%
%% By default, the full list of authors will be used in the page
%% headers. Often, this list is too long, and will overlap
%% other information printed in the page headers. This command allows
%% the author to define a more concise list
%% of authors' names for this purpose.

%%\author{Anonymous authors}
%%\affiliation{
%%	\institution{Paper under double-blind review}
%%}
%%\renewcommand{\shortauthors}{Anonymous Author, et al.}
\author{Wenshuang Liu, Wenting Chen, Linlin Shen$^{*}$}
\affiliation{
	{Computer Vision Institute, College of Computer Science and Software Engineering\\ Guangdong Key Laboratory of Intelligent Information Processing \\Shenzhen University\\
		Email: \{liuwenshuang2018, chenwenting2017\}@email.szu.edu.cn, llshen@szu.edu.cn}}
%%
%% The abstract is a short summary of the work to be presented in the
%% article.
\begin{abstract}
Though GAN (Generative Adversarial Networks) based technique has greatly advanced the performance of image synthesis and face translation, only few works available in literature provide region based style encoding and translation. We propose in this paper a region-wise normalization framework, for region level face translation. While per-region style is encoded using available approach, we build a so called RIN (region-wise normalization) block to individually inject the styles into per-region feature maps and then fuse them for following convolution and upsampling. Both shape and texture of different regions can thus be translated to various target styles. A region matching loss has also been proposed to significantly reduce the inference between regions during the translation process. Extensive experiments on three publicly available datasets, i.e. Morph, RaFD and CelebAMask-HQ, suggest that our approach demonstrate a large improvement over state-of-the-art methods like StarGAN, SEAN and FUNIT. Our approach has further advantages in precise control of the regions to be translated. As a result, region level expression changes and step by step make up can be achieved. The video demo is available at \url{https://youtu.be/ceRqsbzXAfk}.

\end{abstract}

%%
%% The code below is generated by the tool at http://dl.acm.org/ccs.cfm.
%% Please copy and paste the code instead of the example below.
%%


%%
%% Keywords. The author(s) should pick words that accurately describe
%% the work being presented. Separate the keywords with commas.

%% A "teaser" image appears between the author and affiliation
%% information and the body of the document, and typically spans the
%% page.

%%
%% This command processes the author and affiliation and title
%% information and builds the first part of the formatted document.
\maketitle
\begin{figure}[t]
	\centering
	\includegraphics[height=6cm,width=8cm]{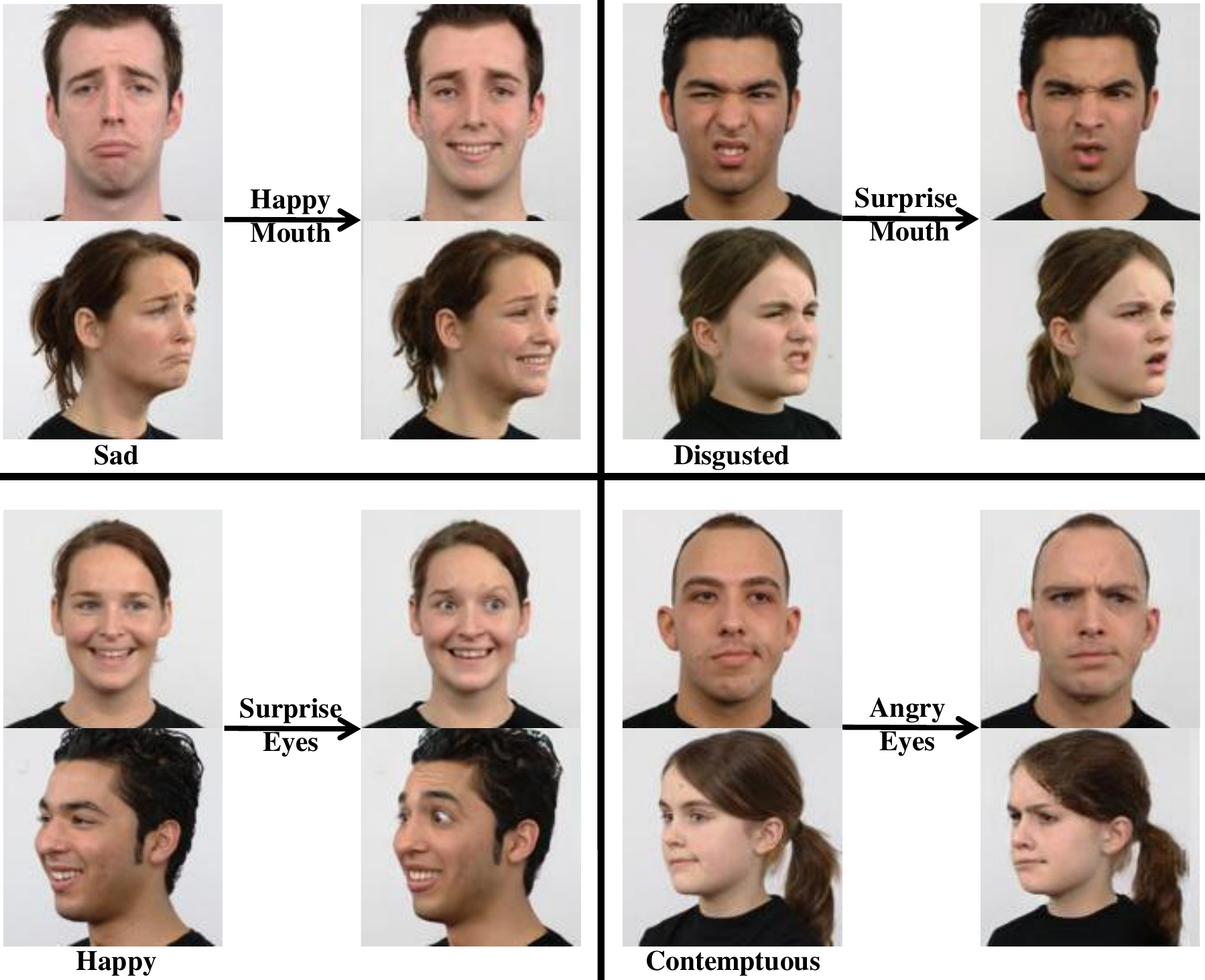}
	
	\caption{Region level face translation: translate the expressions of mouth and eyes for faces in RaFD.}
	\label{rafd_special}
	
\end{figure}

\section{Introduction}
With the development of Generative Adversarial Networks (GANs), the quality of the generated images are getting better. Recent unsupervised image-to-image translation algorithms are remarkably successful in transferring complex appearance changes across different image modalities \cite{DBLP:conf/iccv/ZhuPIE17,DBLP:conf/cvpr/ChoiCKH0C18,DBLP:journals/corr/abs-1912-01865,DBLP:conf/iccv/0001HMKALK19}, we now can use those GANs to generate impressive images. However, current image to image approaches mostly perform translation on the full image, which might not be able to fulfill the requirement of region level translation. For example, we sometime might only want to change the style of a certain part of the faces. As shown in Figure \ref{rafd_special}, while the mouths of faces with sad and disgusted expressions are changed to happy and surprise styles, the eyes of faces with happy and contemptuous expressions are changed to surprise and angry styles. Most of existing image translation approaches cannot achieve this, as they only apply the style changes to the whole image.

Recently, a few conditional Generative Adversarial Networks (cGAN) based attempts, like SPADE (Spatially-adaptive normalization) \cite{DBLP:conf/cvpr/Park0WZ19} and SEAN (Semantic Region-Adaptive Normalization) \cite{DBLP:journals/corr/abs-1911-12861}, have been proposed to synthesize images based on sematic segmentation masks. As SPADE insert style information in beginning of the network, only the same style code can be applied to different semantic regions. To address this issue, SEAN presented an AdaIN \cite{DBLP:conf/iccv/HuangB17} based technique to generate spatially varying normalization parameters and inject these parameters into multiple layers in the network. The styles of different semantic regions can thus be individually encoded and controlled, i.e. various style codes can be applied to different semantic regions. 

\begin{figure}[t]
	\centering
	\includegraphics[height=4.5cm,width=8cm]{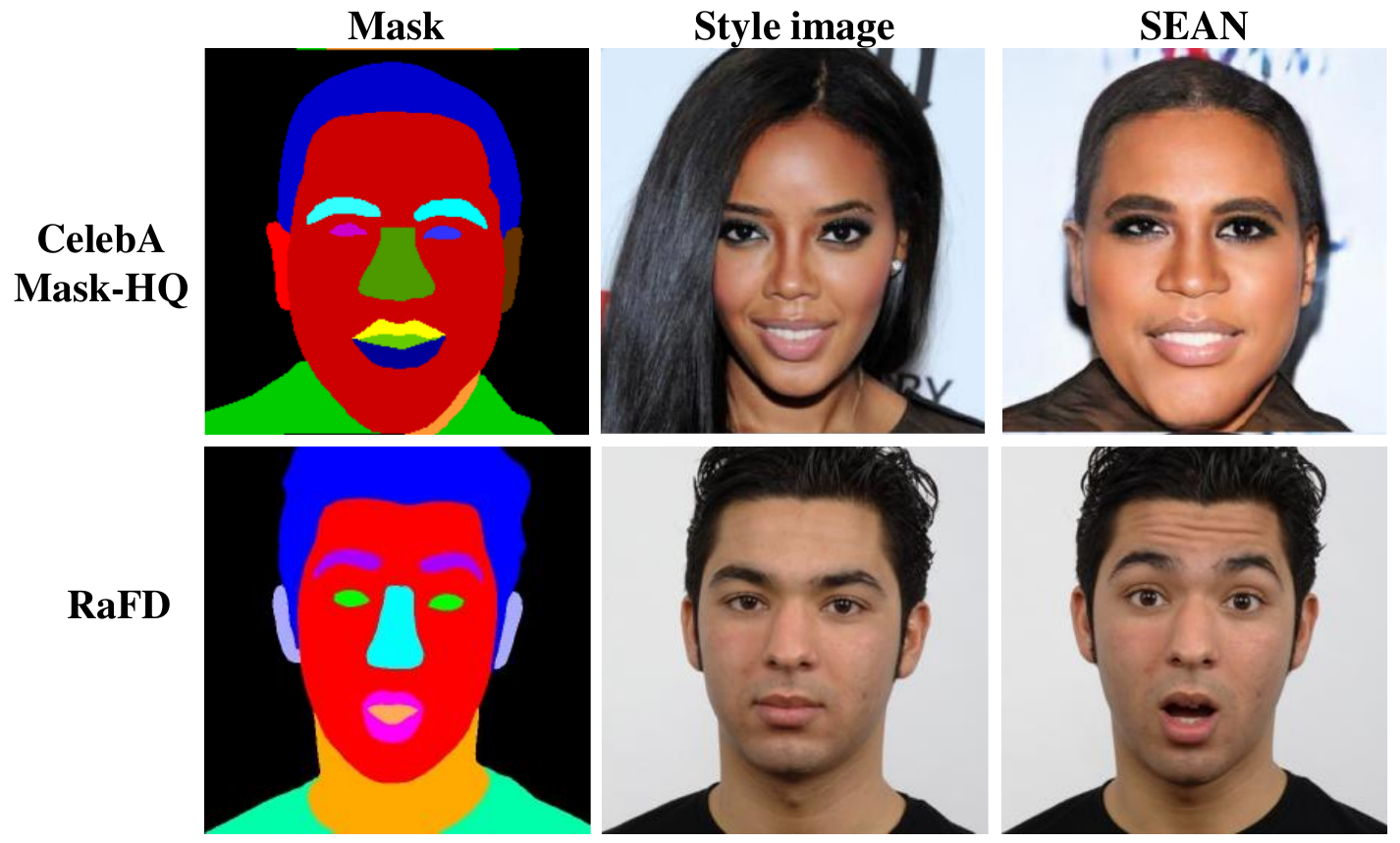}
	\caption{Example faces generated with two different styles by SEAN\cite{DBLP:journals/corr/abs-1911-12861}.}
	\label{sematic_syn}
\end{figure}

However, SEAN is basically designed for image synthesis. Given a segmentation mask, spatially varying styles are used to control the synthesis of different semantic regions. While regions with different styles are synthesized, the shapes of these regions are defined by the segmentation mask and keep fixed during the synthesis. Figure \ref{sematic_syn} shows two example faces synthesized by SEAN from masks with labels of hair, eyes, nose, mouth and so on. The first row shows the face synthesized with reference to the semantic mask of a male face and the style of a lady's face. While skin tone of the synthesized face is similar to that of the style image, the hair of the synthesized face is short, which is pre-defined by the hair region of the input mask. The expression of the face synthesized in the second row, is actually very different with that of the style image. We also did not get good translation results whey trying to translate the eyes and mouths of faces shown in Figure \ref{rafd_special} to different expressions. While region-wise style information (ST) is encoded as a style map and integrated into the blocks of SEAN modules to synthesize faces from the mask M, the feature maps are modulated by the mask and thus the shapes of different regions for the synthesized faces are relatively fixed. To resize the shape of hair, eyes, nose and mouth, an interactive editing tool has to be required to change the contour of corresponding semantic regions in the mask.

To achieve higher degree of freedom in translating different facial regions, we propose in this paper a region-wise normalization block, called RIN, to inject per-region styles into region-wise feature maps, which are then fused together for the following convolution and upsampling process. To reduce the inferences among the translation of different regions, we have also designed a region matching loss to measure the similarity between the regions of content image and style translated image.

We perform extensive experiments on three publicly available datasets, i.e. Morph \cite{DBLP:conf/fgr/RicanekT06}, RaFD \cite{langner2010presentation} and CelebAMask-HQ \cite{DBLP:journals/corr/abs-1907-11922,DBLP:conf/iclr/KarrasALL18,DBLP:conf/iccv/LiuLWT15}. The results are quantitatively evaluated using metrics like Accuracy, FID (Frechet Inception Distance) and LPIPS (Learned Perceptual Image Patch Similarity). Both visual and quantitative results suggest that our approach demonstrate a large improvement over state-of-the-art methods like StarGAN, SEAN and FUNIT. The idea of our work can be summarized as below:

\begin{itemize}	
	\item 	We propose a region based translation framework for face editing. While GAN based approaches usually transfer the style of faces as a whole image, our framework can transfer the style of specified regions, without changing other regions. 
	\item   As our RIN translate the style of content images in a region-wise manner, the introduced building block can generate images more similar to the input style images, by translating both shape and texture of different regions.
	\item   A region matching loss is designed to measure the similarity between translated/non-translated regions, when the style of whole style image, or specific region is applied to transfer a given face. Ablation study shows that our matching loss can significantly reduce the inference between regions during the translation process.

\end{itemize}

\section{Related Work}
\textbf{Generative Adversarial Networks.} Generative Adversarial Networks(GANs) \cite{DBLP:conf/nips/GoodfellowPMXWOCB14} have been successfully applied to various image synthesis tasks, e.g. image inpainting \cite{DBLP:journals/corr/abs-1803-07422,DBLP:conf/cvpr/Yu0YSLH18}, image manipulation \cite{DBLP:conf/iccv/AbdalQW19,DBLP:journals/tog/BauSPWZZ019,DBLP:conf/eccv/ZhuKSE16} and texture synthesis \cite{DBLP:conf/eccv/LiW16,DBLP:conf/eccv/SlossbergSK18,DBLP:journals/tog/FruhstuckAW19}. With continuous improvements on GAN architecture \cite{DBLP:conf/cvpr/KarrasLA19,DBLP:conf/cvpr/Park0WZ19,DBLP:journals/corr/RadfordMC15}, loss function \cite{DBLP:conf/iccv/MaoLXLWS17,DBLP:journals/corr/ArjovskyCB17} and regularization \cite{DBLP:conf/nips/GulrajaniAADC17,DBLP:conf/icml/MeschederGN18,DBLP:conf/iclr/MiyatoKKY18}, the images synthesized by GANs are becoming more and more stable and realistic. For example, WGAN \cite{DBLP:journals/corr/ArjovskyCB17} use Wasserstein distance to regularize the training of GANs, which trys solve the problem of unstable GAN training  and  balance the training process of generator and discriminator. Recently, the human face images generated by StyleGAN V1 \cite{DBLP:conf/cvpr/KarrasLA19}, V2 \cite{DBLP:journals/corr/abs-1912-04958} present very high quality and are almost indistinguishable from photographs by untrained viewers.  A traditional GAN uses noise vectors as the input and thus provides little user control. This motivates the development of conditional GANs (cGANs) \cite{DBLP:journals/corr/MirzaO14} where users can control the synthesis by feeding the generator with conditional information. Examples include class labels \cite{DBLP:conf/iclr/MiyatoK18,DBLP:conf/icml/MeschederGN18,DBLP:conf/iclr/BrockDS19}, text \cite{DBLP:conf/icml/ReedAYLSL16,DBLP:conf/cvpr/HongYCL18,DBLP:conf/cvpr/XuZHZGH018} and images \cite{DBLP:conf/cvpr/IsolaZZE17,DBLP:conf/cvpr/Park0WZ19,DBLP:conf/nips/LiuBK17,DBLP:conf/cvpr/Wang0ZTKC18,DBLP:conf/iccv/ZhuPIE17}. 

\textbf{Image-to-Image Translation.} Image-to-image translation is an umbrella concept that can be used to describe many problems in computer vision and computer graphics. As a milestone, Isola et al. \cite{DBLP:conf/cvpr/IsolaZZE17} first showed that conditional GANs can be used as a general solution to various image-to-image translation problems. Since then, their method is extended by several works to scenarios including unsupervised learning \cite{DBLP:conf/nips/LiuBK17,DBLP:conf/iccv/ZhuPIE17}, few-shot learning \cite{DBLP:conf/iccv/0001HMKALK19}, high resolution image synthesis \cite{DBLP:conf/cvpr/Wang0ZTKC18}, multi-modal image synthesis \cite{DBLP:conf/nips/ZhuZPDEWS17,DBLP:conf/eccv/HuangLBK18} and multi-domain image synthesis \cite{DBLP:conf/cvpr/ChoiCKH0C18,DBLP:journals/corr/abs-1912-01865}. Among various image-to-image translation problems, unsupervised learning and few-shot learning  are particularly useful as they can translate the image into an unseen class image with few images. But they are always translating the whole image, which may not pay enough attention to the details of local regions.

\textbf{Region based style encoding.} 
There have been few image synthesis works in literature related to semantic image generation. Before the proposal of SEAN, SPADE is believed to be the best architecture for semantic image synthesis. However, SPADE uses only one style code to control the entire style of an image.  To allow different styles for different regions in the segmentation masks, SEAN generates spatially varying normalization parameters based on the input of segmentation mask and style images. Per-region style can thus be encoded and applied to different regions. Based on such per-region style encoding, we develop a region-wise normalization block, called RIN, to individually inject per-region styles into region-wise feature maps. The styles of both shape and texture can thus be translated for different regions. A region matching loss has also been proposed to significantly reduce the inference between regions during the translation process.

\begin{figure*}[t]
	\hfill	
	\subfigure[The generator of the proposed RFT.]{
		\centering
		\includegraphics[height=7.5cm,width=16cm]{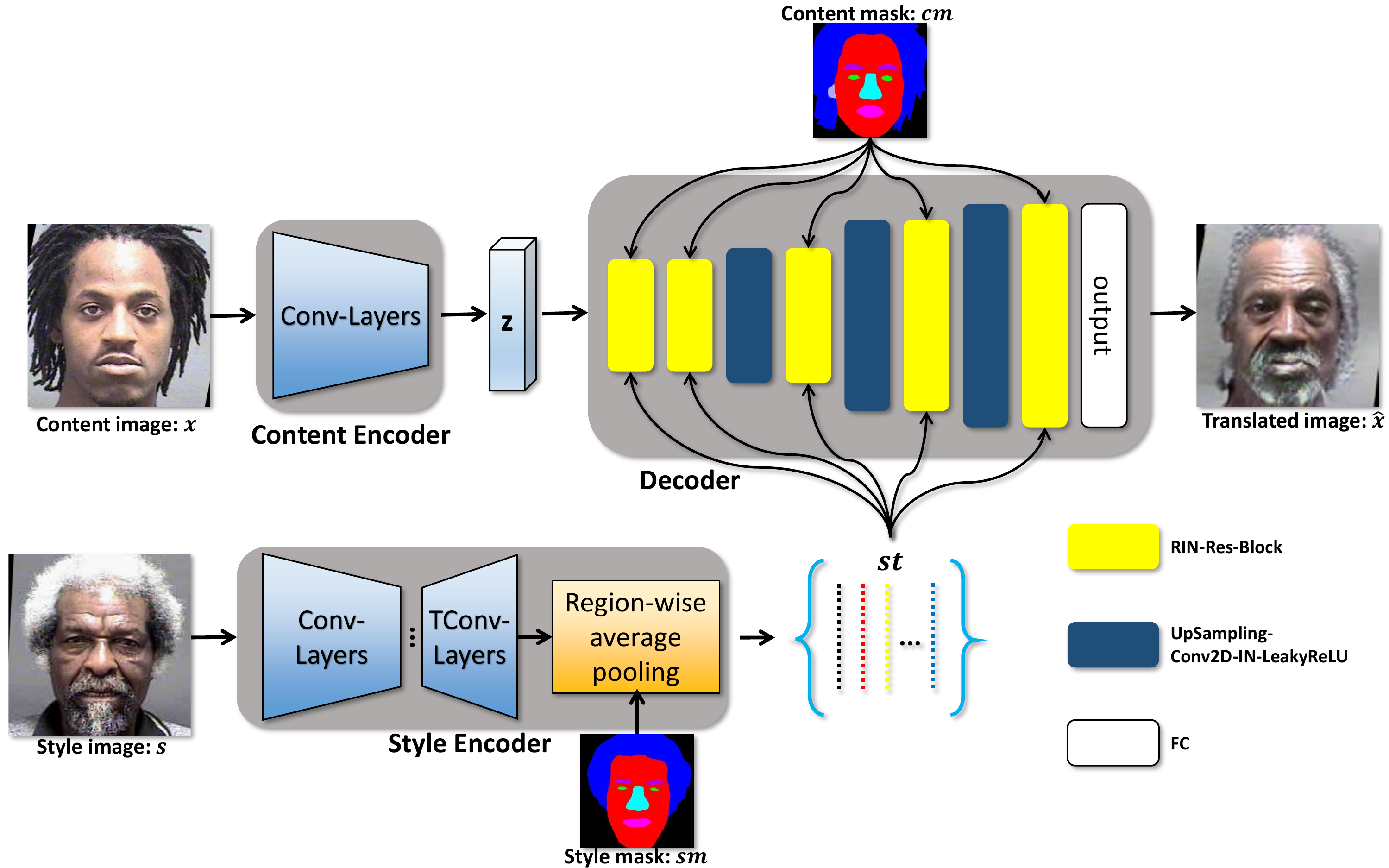}
		\label{network}
	}
	\subfigure[Region-wise Normalization ResBlock.]{
		\begin{minipage}{0.35\textwidth}
			\centering
			\includegraphics[width=0.98\textwidth,height=1.8in]{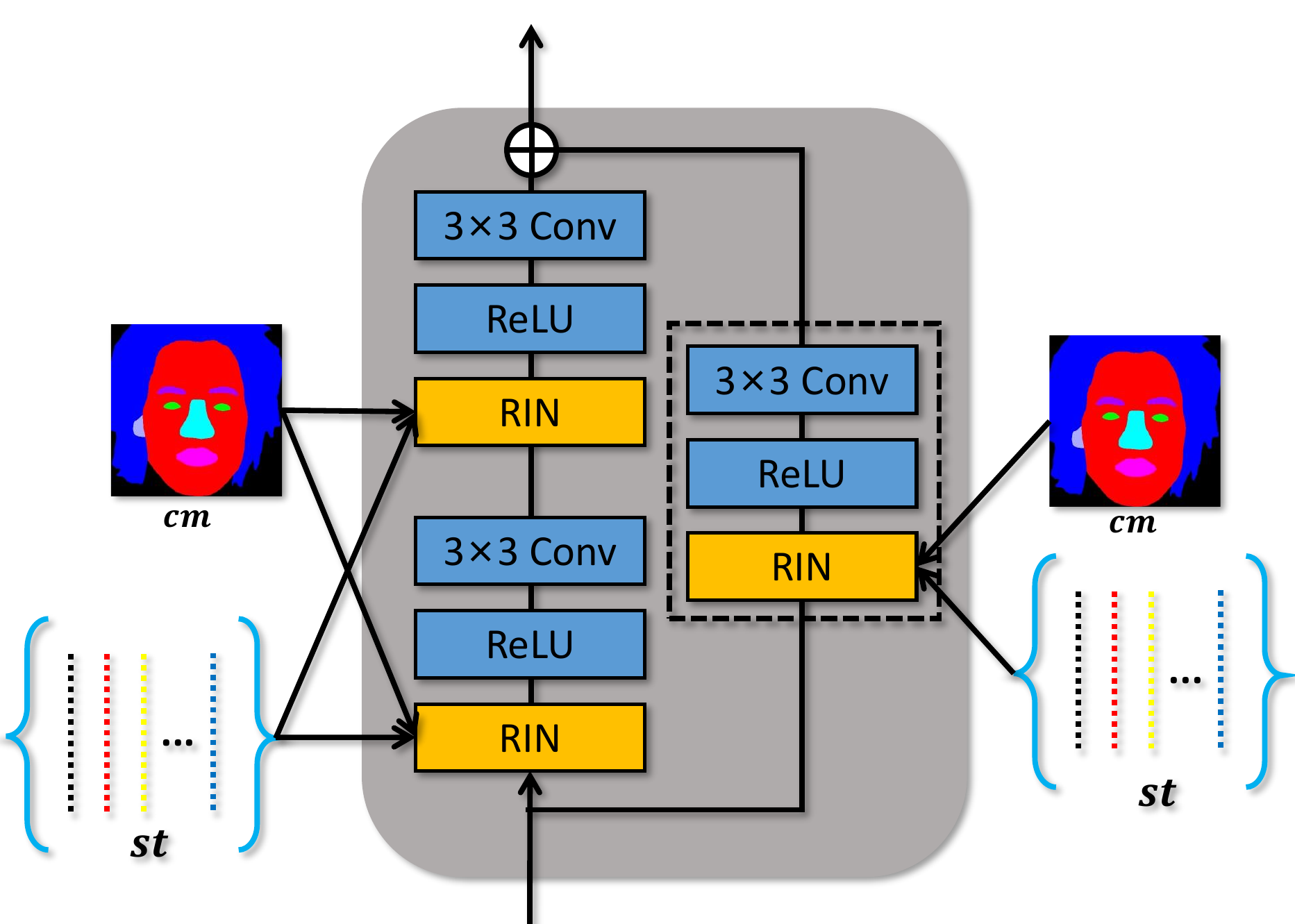}
			\label{RTRNBlock}
		\end{minipage}
	}
	\hfill
	\subfigure[Region-wise Normalization.]{
		\begin{minipage}{0.62\textwidth}
			\centering
			\includegraphics[width=0.98\textwidth,height=2in]{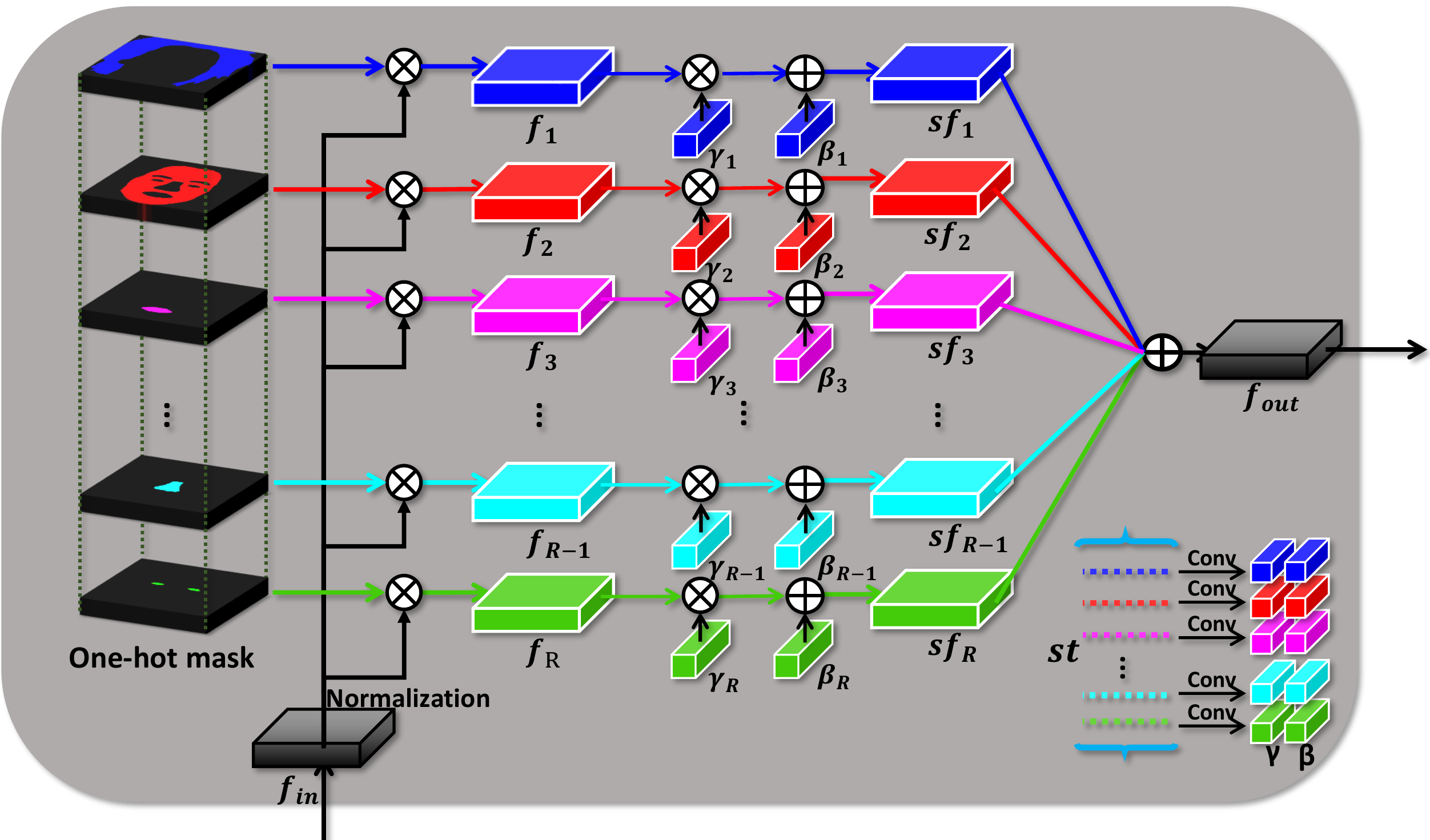}
			\label{R2R}
		\end{minipage}
	}
	\caption{Detailed architecture of our proposed RFT.}
\end{figure*}

\section{Method}
In this paper, we focus on region level face translation. The proposed  framework, named Region-wise Face Translation (RFT), aims to individually translate the styles of different regions in a content image.  As shown in Figure \ref{network}, our generator network architecture consists of content encoder, style encoder and decoder. As shown in the figure, our generator takes four input images (content image $x$, content mask $cm$, style image $s$ and style mask $sm$) and outputs a translated image $\hat{x}$. The process of generation can be represented as:
\begin{equation}
\begin{aligned}
z &= CE(x)          \\
st &= SE(s,sm)           \\
\hat{x} &= De(z,cm,st)
\end{aligned}
\end{equation}
where $CE$ is the content encoder, $SE$ is the style encoder, $De$ is the decoder, $st$ is the style tensor encoded by style encoder, $cm$ is the content mask  with R regions, $\hat{x}$ is the translated image. In the following sections, we will show the details of our architecture.

\subsection{ Style encoder}
As shown in Figure \ref{network}, inspired by SEAN, our encoder employs a bottleneck structure to remove the information irrelevant to styles from the style image. The feature map extracted by $TConv-Layers$ (transposed convolution) will be passed through a Region-wise average pooling module to get style tensor $st$. Each vector in $st$ corresponds to one region in style mask. In implementation, we first transform style mask into one-hot tensor where each channel represents a region. Take a channel representing hair region for example, while the values of pixels in hair region are set as 1, others are set to 0. A set of R style feature maps can then be obtained by element-wise multiplication between feature map and different one-hot channels. Finally, we use a global average pooling to get style tensor $st$, which consists of style information of the R regions.

\subsection{ Decoder}
As shown in Figure \ref{network}, the decoder is composed of five $RIN-Res$ blocks, three $UpSampling$ blocks and one $Fully-Connected$ layer. As shown in Figure \ref{RTRNBlock}, our proposed $RIN-Res$ block consists of three convolutional layers, three $ReLU$ layers and three $RIN$ blocks.  Each $RIN$ residual block takes three inputs: content feature maps, per-region style tensor $st$ and content mask. Note that the input content mask is downsampled to the same height and width of the feature maps at the beginning of each $RIN$ block.

\begin{figure*}[t]
	\centering
	\includegraphics[height=5cm,width=16cm]{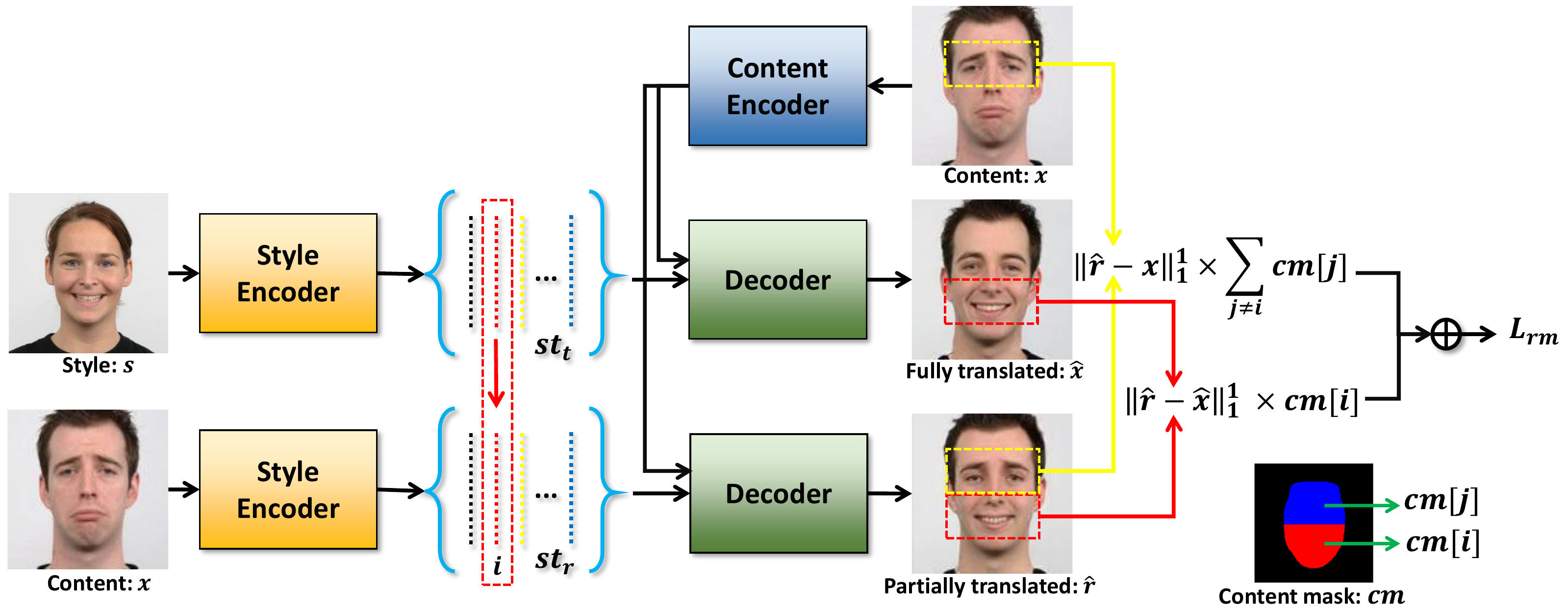}
	\caption{ Region matching loss.}
	\label{RMLoss}
\end{figure*}
\subsubsection{  Region-wise normalization}
Given a style tensor $st$ encoding R region styles, the segmentation mask of content image $cm$ and input feature map $f_{in}$, our $RIN$ block tries to translate each of the R regions in content image to the corresponding style specified in the tensor $st$, by region-wise normalization. As shown in Figure 4(c), we fist multiply (element-wise) feature map $f_{in}$ with the one-hot masks (channels) to get per-region feature maps $\{f_1,f_2,…,f_R\}$, which are then modulated by the normalization parameters learned from the style tensor $st$.   Let $f_{in}$ denote the input feature map of the current $RIN$ block in a deep convolutional network for a batch of $N$ samples, $H$, $W$ and $C$ be the height, width and channel numbers of the feature map, the style feature map of the $ith$ region at site $f_{i}^{n,c,y,x}$ ($n\in N, c\in C, y\in H, x\in W$) can be represented as:
\begin{equation}
f_{i}^{n,c,y,x} = \frac{f_{in}^{n,c,y,x}- \mu^{c}}{\sigma^{c}} \times cm[i]
\end{equation}
where $f_{in}^{n,c,y,x}$ denote the feature map at the site before normalization,  $cm[i]$ denotes the one-hot mask corresponding to the $ith$ region, $\mu^{c}$ and $\sigma^{c}$ are the mean and standard deviation of the feature map in channel $c$:
\begin{equation}
\mu^{c}= \frac{1}{NHW}\sum_{n,y,x}f_{in}^{n,c,y,x}
\end{equation}

\begin{equation}
\sigma^{c}= \sqrt{\frac{1}{NHW}\sum_{n,y,x}(f_{in}^{n,c,y,x})^2 - (\mu^{c})^2}
\end{equation}

After getting the per-region feature map of content image, with the same operation as AdaIN \cite{DBLP:conf/iccv/HuangB17}, we do the element-wise calculation between the per-region feature map and its corresponding regional modulation parameters $\gamma$ and $\beta$ extracted by $st$:
\begin{equation}
sf_{i}^{n,c,y,x} =f_{i}^{n,c,y,x}\times(1 + \gamma_i )+ \beta_i
\end{equation}
where $sf_{i}^{n,c,y,x}$ denotes style feature map for the $ith$ region, $\gamma_i$ and $\beta_i$ are  the modulation parameters learned from the $ith$ channel of $st$. 

By now, the per-region feature maps have been all injected with per-region styles encoded from style image, using our region-wise normalization. Finally, the R modulated per-region feature maps are added together to get the  output feature map:
\begin{equation}
f_{out}^{n,c,y,x} = \sum_{i}sf_{i}^{n,c,y,x}
\end{equation}

\subsection{Discriminator}
The discriminator architecture of RFT is the same as that of FUNIT \cite{DBLP:conf/iccv/0001HMKALK19}. As our RFT aims to translate the styles of specified regions only, we proposed a novel region matching loss to reduce the interferences among different regions.

\subsubsection{ Region Matching Loss}
As shown in Figure \ref{RMLoss}, we first use a content image $x$ and a style image $s$ to generate a face $\hat{x}$ presenting similar expression with $s$, which can be represented as:
\begin{equation}
\begin{aligned}
st_t &= SE(s,sm) \\
\hat{x} &= De(CE(x),st_t,cm)
\end{aligned}
\end{equation}
where $st_t$ is the style tensor encoded from the R regions of style image $s$ and $\hat{x}$ is the result image where all R regions have been translated to the per-region styles encoded in $st_t$. In the second task, we only translate the style of $ith$ region of the content image $x$, by replacing the $ith$ channel of its style tensor, with that of $st_t$:
\begin{equation}
\begin{aligned}
st_r &= SE(x,cm) \\
st_r&[i] = st_t[i] \\
\hat{r} = &De(CE(x),st_r,cm)
\end{aligned}
\end{equation}
where $st_r[i]$ and  $st_t[i]$ are the $ith$ channel of style tensor $st_r$ and $st_t$, respectively, which encode the style of the $ith$ region of $x$ and $s$, $\hat{r}$ represent the result image by translating the style of $ith$ region of content image $x$.

Given a content image $x$ and the fully translated image $\hat{x}$ and partially translated $\hat{r}$, we design a region matching loss to measure the similarity between the $ith$ regions of  $\hat{x}$ and $\hat{r}$, and the similarity between other regions of $x$ and $\hat{r}$: 
\begin{equation}
\mathcal{L}_{RM} = E_{x,\hat{r},\hat{x}}[||\hat{r}-\hat{x}||_1^1 \times cm[i] + ||\hat{r}-x||_1^1 \times \sum_{j \neq i} cm[j]]
\end{equation}
where $cm[i]$ and $cm[j]$ represent the one-hot mask corresponding to the $ith$ and $jth$ regions, respectively.
\subsection{Training}
The proposed RFT was trained  by solving a minimax optimization problem given by
\begin{equation}
\label{GAN_loss}
\begin{array}{cc}
\mathop{min}\limits_{D}\mathop{max}\limits_{G}\mathcal{L}_{GAN}(D,G)+ \lambda_R\mathcal{L}_R(G)+\qquad\qquad\\\qquad\qquad\qquad\qquad
\lambda_{FM}\mathcal{L}_{FM}(G)+\lambda_{RM}\mathcal{L}_{RM}(G)
\end{array}
\end{equation}
where $\mathcal{L}_{GAN}$, $\mathcal{L}_R$, $\mathcal{L}_{FM}$ and $\mathcal{L}_{RM}$ are the GAN loss, the content image reconstruction loss, the feature matching loss and the region matching loss, respectively. The GAN loss is a conditional one given by
\begin{equation}
\begin{array}{cc}
\mathcal{L}_{GAN}(D,G) = E_x[-logD^{c_x}(x)]+\qquad\qquad\\\qquad\qquad\qquad\qquad
E_{x,\{y_1,...,y_k\}}[log(1-D^{c_y}(\hat{x})]
\end{array}
\end{equation}

The content reconstruction loss helps G learn a translation model. Specifically, when using the same image for both the input content image and the input style image, the loss encourages G to generate an output image identical to the input
\begin{equation}
\begin{array}{cc}
\mathcal{L}_R(G) = E_x[||x-G(x,cm,\{x,cm\})||_1^1]
\end{array}
\end{equation}

The feature matching loss regularizes the training. We first construct a feature extractor, referred to as $D_f$, by removing the last (prediction) layer from $D$. We then use $D_f$ to extract features from the translation output $\hat{x}$ and the style image $\{y_1,...,y_k\}$ and minimize
\begin{equation}
\mathcal{L}_{FM}(G) = E_{x,\{y_1,...,y_k\}}[||D_f(\hat{x})-\sum_{k}\frac{D_f(y_k)}{k}||_1^1]
\end{equation}

The GAN loss, the content reconstruction loss and the feature matching loss are the same as that of FUNIT.

\section{Experiment}
Our proposed RFT was evaluated on three challenging datasets, i.e. Morph, RaFD, CelebAMaskHQ. A wide range of quantitative metrics including FID, Accuracy and LPIPS were evaluated among different models; Qualitatively, the examples of synthesized images are shown for visual inspection.
\subsection{Dataset}
\textbf{Morph.} Morph dataset \cite{DBLP:conf/fgr/RicanekT06} is a large-scale public longitudinal face dataset, collected in indoor office environment with variations in age, pose, expression and lighting conditions. It has two subsets: Album1 and Album2. Album 2 contains 55,134 images of 13,000 individuals with age label ranging from 16 to 77 years old. We divide the images into a training set with 50020 images and a test set with 4,925 images. The images are separated into five groups with ages of 11-20, 21-30, 31-40, 41-50 and 50+.

\textbf{RaFD.} RaFD dataset \cite{langner2010presentation} is a high-quality face database, containing a total of 67 models with 8,040 pictures displaying 8 emotional expressions, i.e., angry, fearful, disgusted, contempt, happy, surprise, sad and neutral. Each expression consists of three different gaze directions and was simultaneously photographed from different angles using five cameras.  We divide the images into a training set with 4,320 images and a test set with 504 images.

\textbf{CelebAMask-HQ.}  CelebAMask-HQ dataset \cite{DBLP:journals/corr/abs-1907-11922,DBLP:conf/iclr/KarrasALL18,DBLP:conf/iccv/LiuLWT15} containing 30,000 segmentation masks for the CelebAHQ face image dataset. There are 19 different region categories in CelebAMask-HQ dataset. We divide the images into a training set with 25,000 images and a test set with 5,000 images.
\begin{figure}[t]
	\centering
	\includegraphics[height=4.8cm,width=5cm]{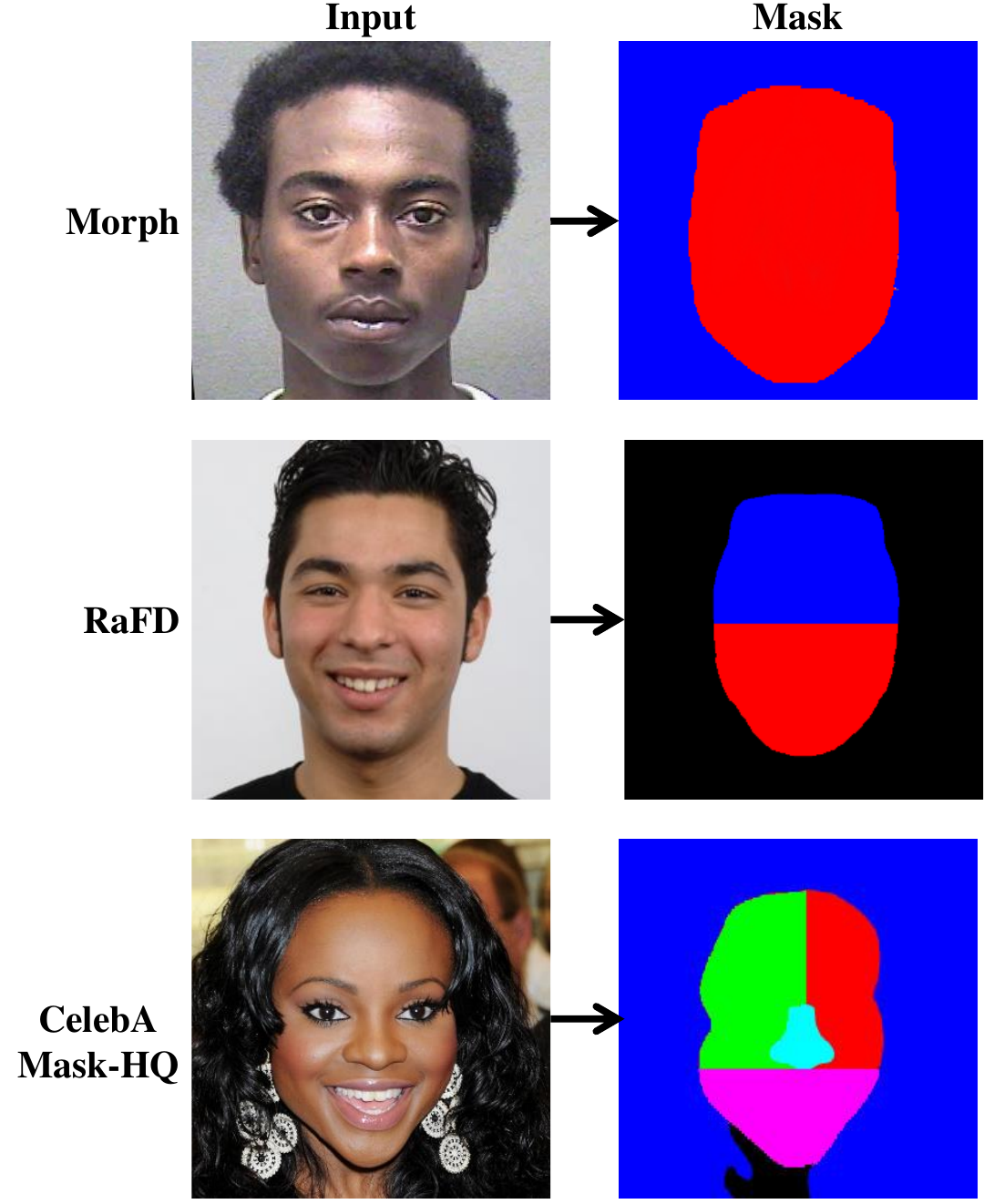}
	
	\caption{Different masks used for Morph, RaFD and CelebAMask-HQ.}
	\label{Mask_setting}
\end{figure}
\subsection{Metrics}
In the training stage of three datasets above, we train different GAN models with their training set. Note that all the baselines are trained with the batch size of 4, the image size of $128\times128$ and  the maximum iteration of 100,000. As show in Figure \ref{Mask_setting}, for variant translation tasks, we design different mask setting for different datasets.

In the test stage, we evaluate performance of different models on their test set using three metrics as follows:

\textbf{Accuracy.} Three classifiers (Resnet-18) \cite{DBLP:conf/cvpr/HeZRS16} trained using three training sets of different datasets are used to test accuracy of translation. If the synthetic face of target class is correctly classified by the classifier, we decide such translation as a successful one.

\textbf{FID.} Calculated as the Frechet inception distance \cite{DBLP:conf/nips/HeuselRUNH17} between two feature distribution of the generated and real images, FID score has been shown to correlate well with human judgement of visual quality. It measures the similarity between two sets of images. Lower FID value indicates better quality of the synthetic images.   We use the ImageNet-pretrained Inception-V3 \cite{DBLP:conf/cvpr/SzegedyVISW16} classifiers as feature extractor. For each test image from a source domain, we translate it into a target domain using 10 style images  randomly sampled from the test set of the target domain. We then compute the FID between the translated images and training images in the target domain. We compute the FIDs for every pair of image domains and report the average score.

\textbf{LPIPS.} Learned perceptual image patch similarity (LPIPS) \cite{DBLP:conf/cvpr/ZhangIESW18} measures the diversity of the generated images using the L1 distance between features extracted from the pretrained AlexNet \cite{DBLP:journals/cacm/KrizhevskySH17}. For each test image, we translate its style with reference to 10 style images randomly sampled from the target domain. The L1 distances between each pair of translated image and the style image are then averaged as the LPIPS of the test image.  Finally, we report the average of the LPIPS values over all test images. Note that LPIPS is not available for StarGAN, as it does not require any style image for face translation.

\begin{figure*}[t]
	\centering
	\includegraphics[height=5.6cm,width=15cm]{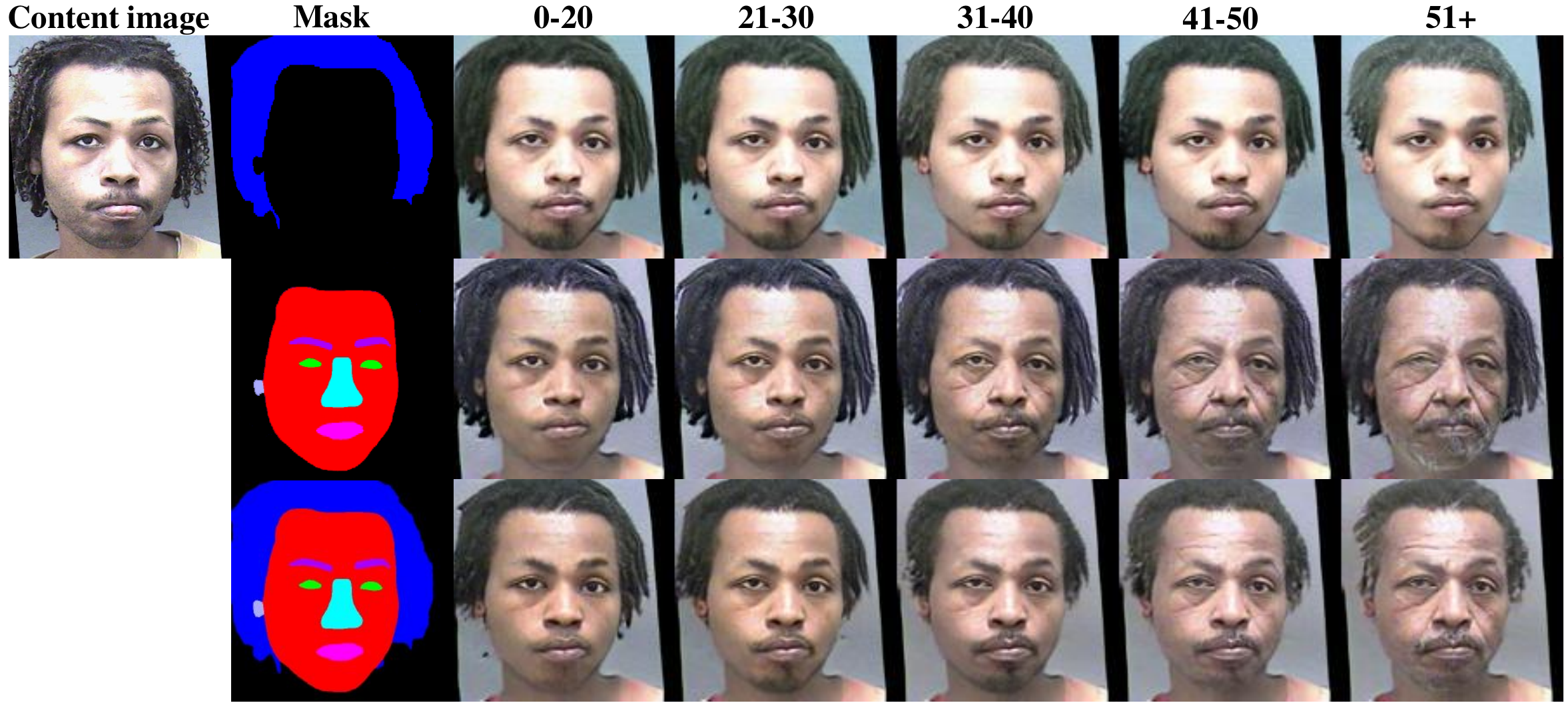}
	\caption{ Translate hair or/and face into styles of different age groups for an example face in Morph.}
	\label{Morph_region}
\end{figure*}

\begin{figure*}[t]
	\centering
	\includegraphics[height=5.6cm,width=17cm]{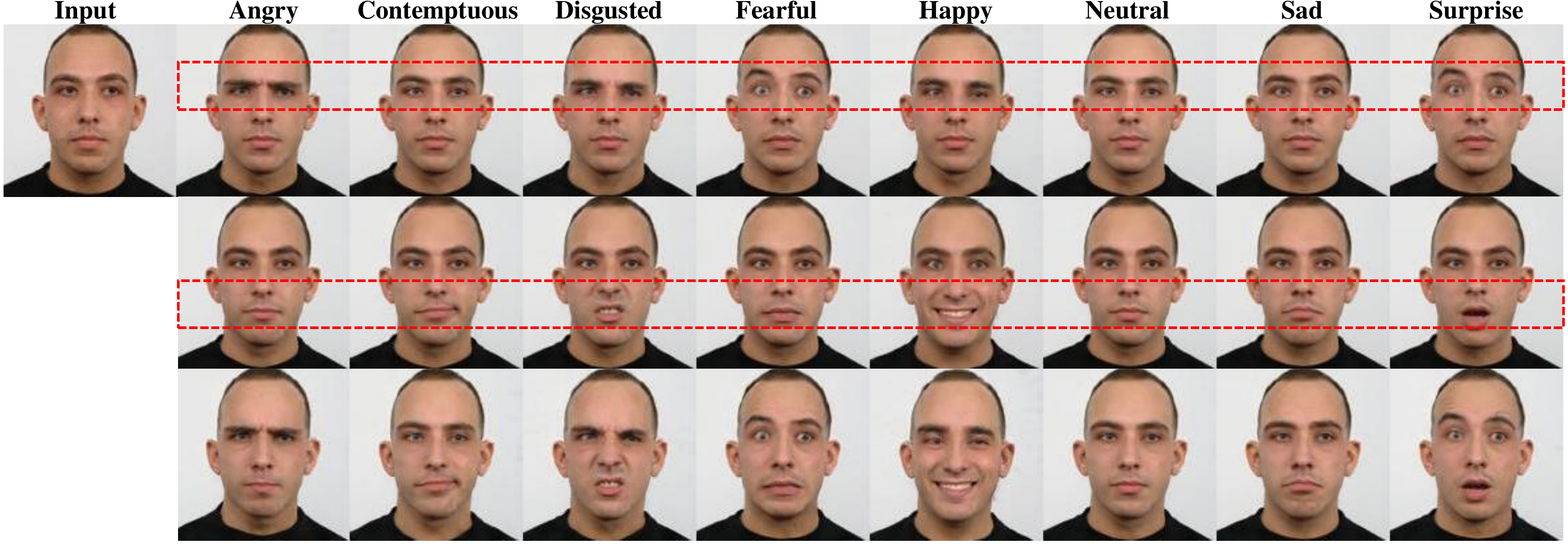}
	
	\caption{Translate eyes or/and mouth into different expressions for an example face in RaFD.}
	\label{RAFD}
	
\end{figure*}
\subsection{Results on Morph}
Firstly, the RFT is evaluated on Morph dataset to assess region level age attribute translation. Figure \ref{Morph_region}  shows the results of our RFT. Note that the per-region styles are encoded using 10 style images randomly sampled from the test set of the target age groups. Figure 7 shows the translation results of an example face of a 25 years old man. In the first row, the hair of the young man is translated to the styles of different age groups (long black to short white), with fixed face regions. In the second row, the face of the young man is translated to the styles of different age groups (appearance of wrinkles), with fixed hair style. In the third row, both hair and face are translated. One can visually observe that our RFT can well control the regions to be translated and achieve decent styles for target regions.

\begin{table}[t]
	\caption{Results of the GAN based models on Morph.}
	\centering
	\begin{tabular}{cccc}
		\toprule
		Method & Accuracy(\%)  & FID score & LPIPS\\
		\midrule
		StarGAN  &60.88  & 27.89  & - \\
		SEAN  &  30.25    & 48.84   & 0.2525\\
		FUNIT  & 39.02    & 26.14   & 0.3152\\
		\textbf{RFT}  &\textbf{69.01}    & \textbf{23.34}  & \textbf{0.2512} \\
		\bottomrule       
	\end{tabular}
	\label{Morph_quality}
\end{table}
The accuracy, FID and LPIPS of face images translated by our RFT are listed in Table \ref{Morph_quality}, together with that translated by StarGAN, SEAN and FUNIT. One can observe from the table that the accuracy of RFT is as high as 69 $\%$, which is significantly higher than that of StarGAN, SEAN and FUNIT. Also, our method achieves the lowest FID score and LPIPS among these GAN based models.

In addition to region level translation, our RFT also does well for image level translation. Figure \ref{ref_morph} in Appendix shows five example faces translated to different age groups specified by the style images listed in the first row. While clear hair changes and wrinkles can be observed, the identity of the face is well preserved.

\begin{figure*}[t]
	\centering
	\includegraphics[height=8.5cm,width=16cm]{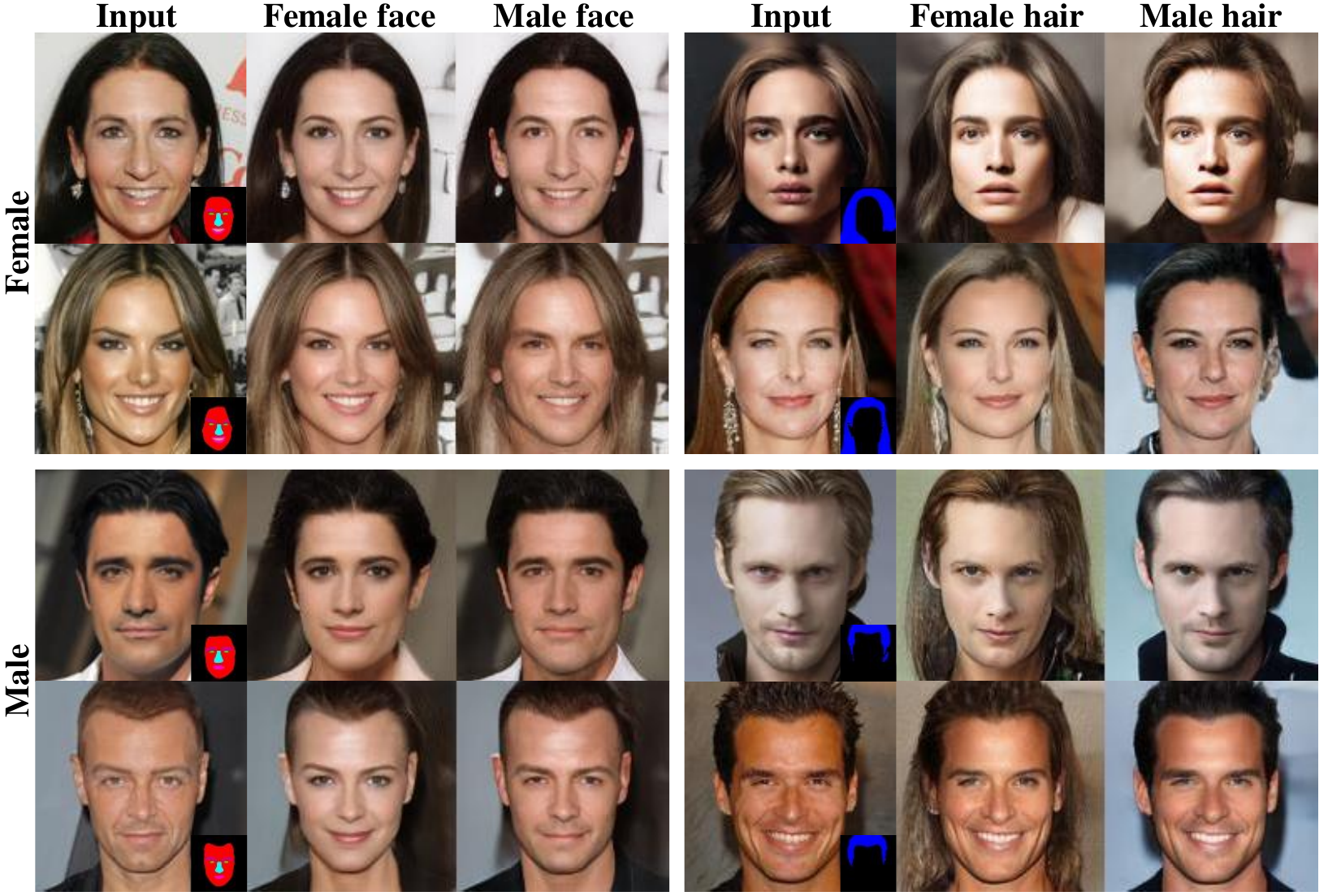}
	\caption{Translate the hair/faces of example faces in CelebAMask-HQ to different genders.}
	\label{CelebA}
\end{figure*}
\begin{figure*}[t]
	\centering
	\includegraphics[height=5cm,width=17cm]{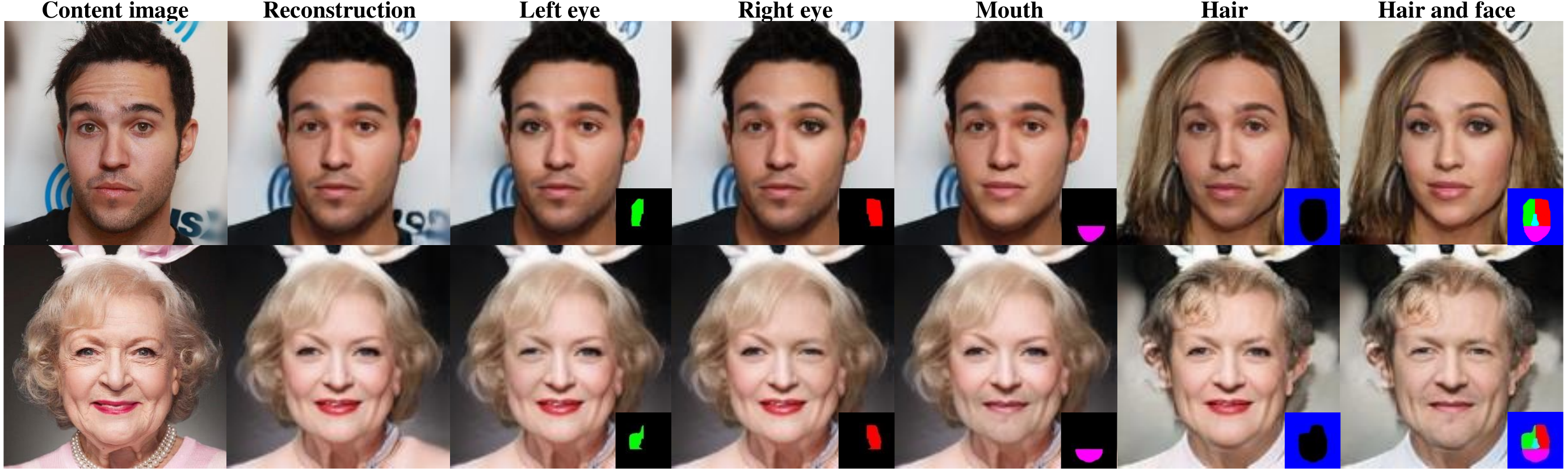}
	
	\caption{Translate the gender styles of left/right eyes, mouth, hairs and faces of two example faces in CelebAMask-HQ.}
	\label{Celeba_R_add}
	
\end{figure*}
\subsection{Results on RaFD}
We now test the performance of our RFT for region level expression translation using RaFD dataset. Figure \ref{RAFD} shows the translation result of an example face in RaFD, whose eyes or/and mouth are translated from neutral to different expressions like angry, fearful, happy, sad and surprise, etc. In the first and second rows, only the eyes and mouth of the man are respectively translated to different expressions, with other regions fixed. In the third row, both eyes and mouth are translated. One can observe from the figure that our approach can precisely translate the shape and texture of designated facial regions to a target expression, without touching any other regions.

\begin{table}[t]
	
	\caption{Results of the GAN based models on RaFD.}
	\centering
	\begin{tabular}{cccc}
		\toprule
		Method & Accuracy(\%)  & FID score & LPIPS\\
		\midrule
		StarGAN  & 77.28  & 32.67  & - \\
		SEAN  & 13.10    &29.61   &  \textbf{0.2610} \\
		FUNIT  & 12.72    &41.67   & 0.2937\\
		\textbf{RFT}  &\textbf{88.32}    & \textbf{27.88}  & 0.2776 \\
		\bottomrule       
	\end{tabular}
	\label{Rafd_quality}
\end{table}
Table \ref{Rafd_quality} shows the accuracy, FID and LPIPS of faces generated by different GAN models. One can observe from the table that the accuracy of RFT is as high as 88.32 $\%$, which is significantly higher than that of StarGAN and more than 75$\%$ higher than that of SEAN and FUNIT. Also, our method achieves the lowest FID of $27.88$, which is $13.79$ lower than that of FUNIT. Though the FID of SEAN is close to our RFT, the expressions translated by SEAN is not accurate, due to the fixed shape defined in the semantic mask (see Figure \ref{sematic_syn} for an example). Figure \ref{FUNIT_RAFD} in Appendix presents more example faces with different expressions translated by StarGAN, SEAN, FUNIT and our RFT, which clearly justify the advantages of our approach, in terms of the visual quality of generated face images.

\begin{table}[t]
	
	\caption{Results of the GAN based models on CelebAMask-HQ.}
	\centering
	\begin{tabular}{cccc}
		\toprule
		Method & Accuracy(\%)  & FID score & LPIPS\\
		\midrule
		StarGAN & 62.17 & 47.53   &-\\
		MUNIT  & 81.65  & 37.07  & 0.4155 \\
		SEAN   &  72.95   & 61.06   &0.3465 \\
		FUNIT  & 93.30    & 35.17   & 0.3781 \\
		\textbf{RFT}  &\textbf{97.06}    & \textbf{31.06}  & \textbf{0.3450} \\
		\bottomrule       
	\end{tabular}
	\label{CelebA_quality}
\end{table}

\begin{figure*}[htp]
	\centering
	\includegraphics[height=9cm,width=17cm]{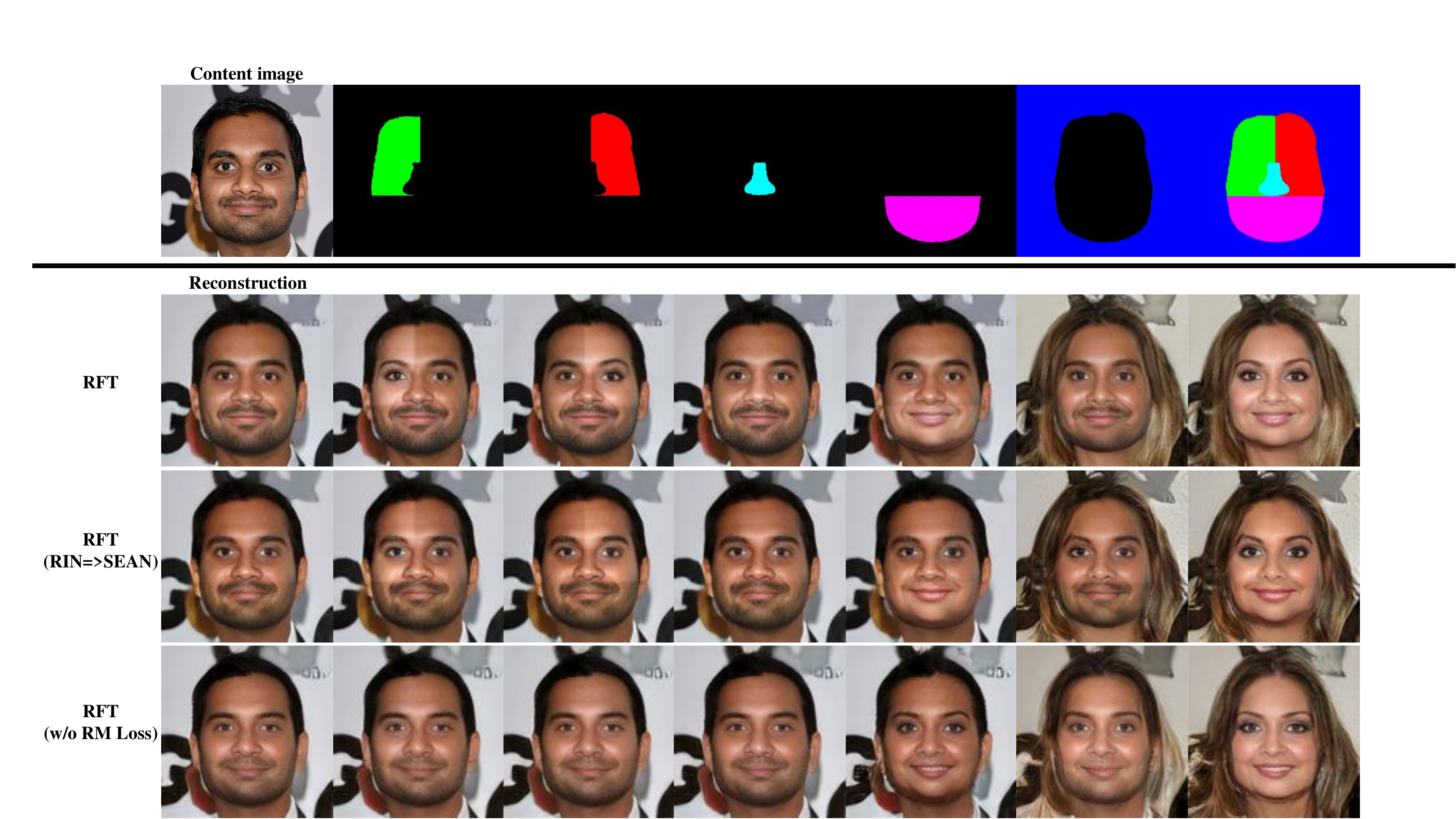}
	\caption{Result for Ablation studies.}
	\label{ablation_img}
\end{figure*}
\subsection{Results on CelebAMask-HQ}
We now evaluate the region level gender translation performance of our approach using CelebAMask-HQ dataset. Figure \ref{CelebA} shows the results of translating face or hair of example faces to different genders. While the faces in the left column are translated to the style of opposite genders, with fixed hair style, the hair styles of the faces in the right column are translated to long/short, with fixed facial styles. Figure \ref{Celeba_R_add} further shows the results of a man and lady when their left/right eyes, mouths, hairs and full images are translated to the styles of opposite genders, using the mask presented in Figure \ref{Mask_setting}. Again one can observe that our model can precisely translate the style of region controlled by the mask overlaid on the bottom right corner of the generated faces, without touching other regions.

Table \ref{CelebA_quality} lists the accuracy, FID and LPIPS of different approaches. Again, our RFT achieves the highest accuracy (97.06$\%$) and lowest FID (31.06) and LPIPS (0.3450).

Figure \ref{makeup} in Appendix shows the translation of left/right eyes, nose, mouth and faces to the style of a beautiful lady.  When the five regions are translated one by one, one can clearly see the make up effects like eye-shadow and whitening of the skin, which beautify the faces to make the ladies look more attractive.

\begin{table}[t]
	
	\caption{Result for Ablation studies.}
	\centering
	\begin{tabular}{cccc}
		\toprule
		Method & Accuracy(\%)  & FID score & LPIPS\\
		\midrule
		RFT(RIN$\Rightarrow$SEAN)  & 96.85 & 35.01  &0.3494 \\
		RFT(w/o RM loss)  & 96.61    &33.81   & \textbf{0.3421}\\
		\textbf{RFT}  &\textbf{97.06}    & \textbf{31.06}  & 0.3450 \\
		\bottomrule       
	\end{tabular}
	\label{ablation}
\end{table} 

\subsection{Ablation studies on CelebAMask-HQ}
To further prove the effectiveness of our proposed RIN block and RM (region matching) loss, we perform an ablation study in this section. We replaced our RIN block with SEAN, removed the RM loss, i.e. set $\lambda_{RM} =0$ in equation (\ref{GAN_loss}), and tested the performance of RFT for gender style translation using CelebAMask-HQ dataset. Figure 11 shows the translation results of different regions for a young man when RFT with different settings are applied. Compared with RFT using SEAN blocks, the left/right eye and nose (the 2nd, 3rd and 4th columns) translated by the original RFT present more lady-like styles, i.e. eye-shadows appear around the eyes and the nose is whitened. When RM is removed, there is no significant difference among the faces presented in the third row when left/right eye and nose are translated, respectively. The long hair in the sixth column actually does not fit the face boundary well. Table \ref{ablation} lists the accuracy, FID and LPIPS of different settings. Compared with SEAN, our RIN block significantly reduces FID from 35.01 to 31.06. The accuracy of our RFT is also higher than that with SEAN and trained without RM loss.

\section{conclusion}
This paper proposed a novel region-wise face translation network, named RFT, region based face translation. A region-wise normalization block and region matching loss are proposed to fuse the per-region style of style and content images, and reduce the influence between different regions, respectively. The proposed RFT is evaluated on three datasets and the experiments results demonstrates its effectiveness.
%%
%% The next two lines define the bibliography style to be used, and
%% the bibliography file.
\bibliographystyle{ACM-Reference-Format}
\bibliography{ACMMM}

%%
%% If your work has an appendix, this is the place to put it.

\appendix
\clearpage
\section{Appendix}
\begin{figure*}[b]
	\centering
	\includegraphics[height=13cm,width=15cm]{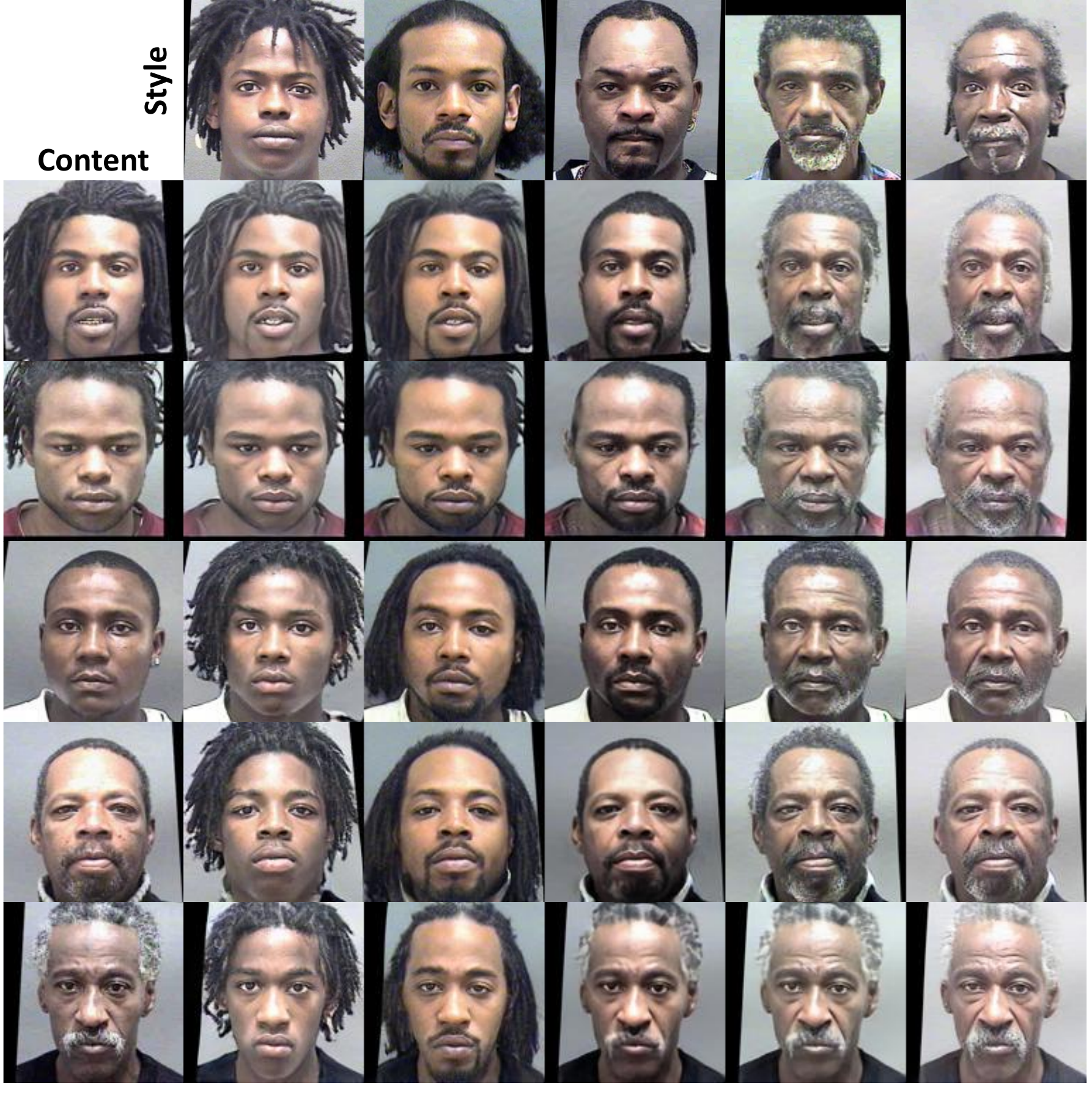}
	\caption{Translation results of our RFT model for more example faces in Morph.}
	\label{ref_morph}
\end{figure*}
\begin{figure*}[p]
	\centering
	\includegraphics[height=12cm,width=16cm]{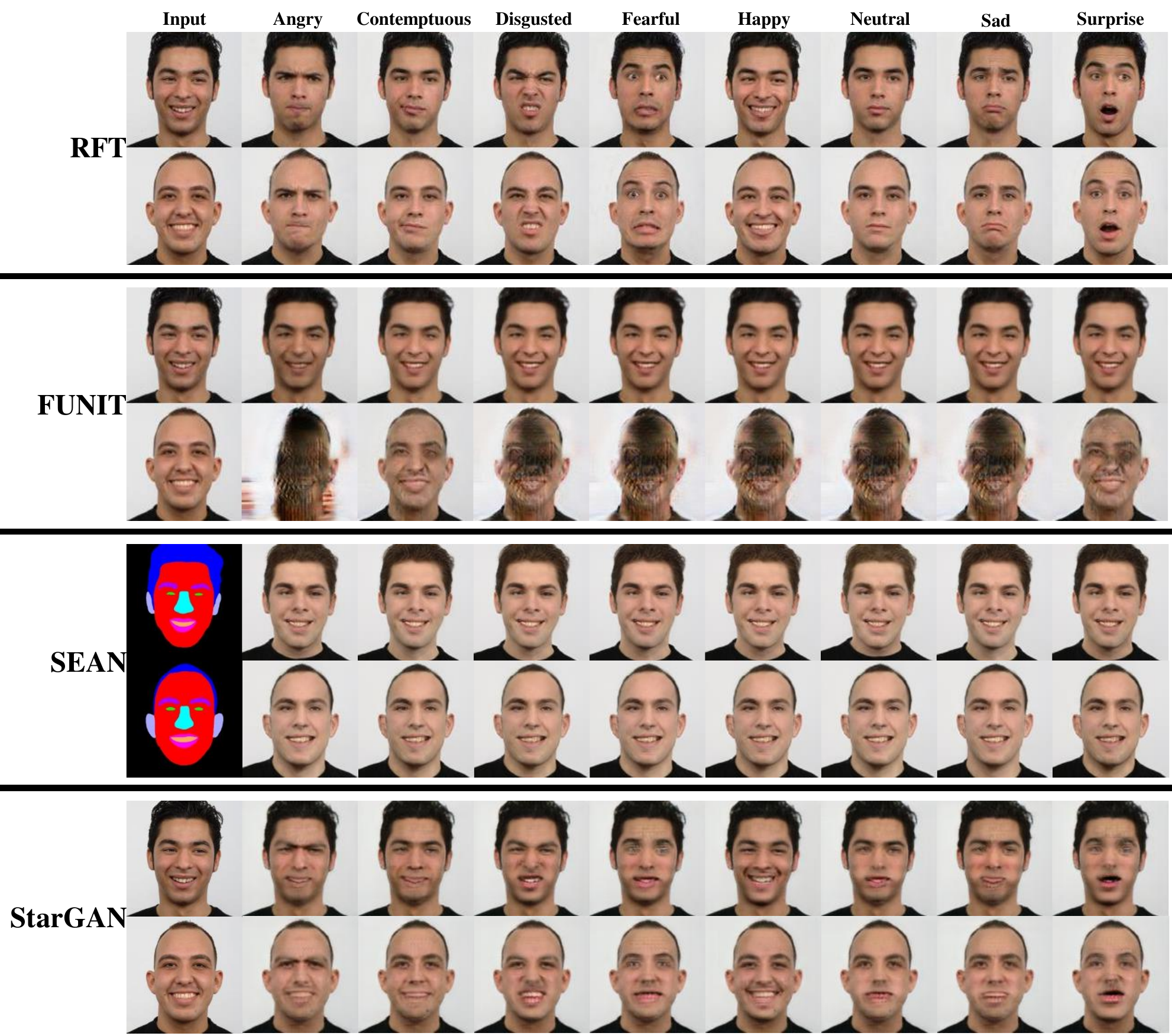}
	\caption{Faces with different expressions translated by  RFT, FUNIT, SEAN and StarGAN on RaFD.}
	\label{FUNIT_RAFD}
\end{figure*}

\begin{figure*}[htp]
	\centering
	\includegraphics[height=20cm,width=14cm]{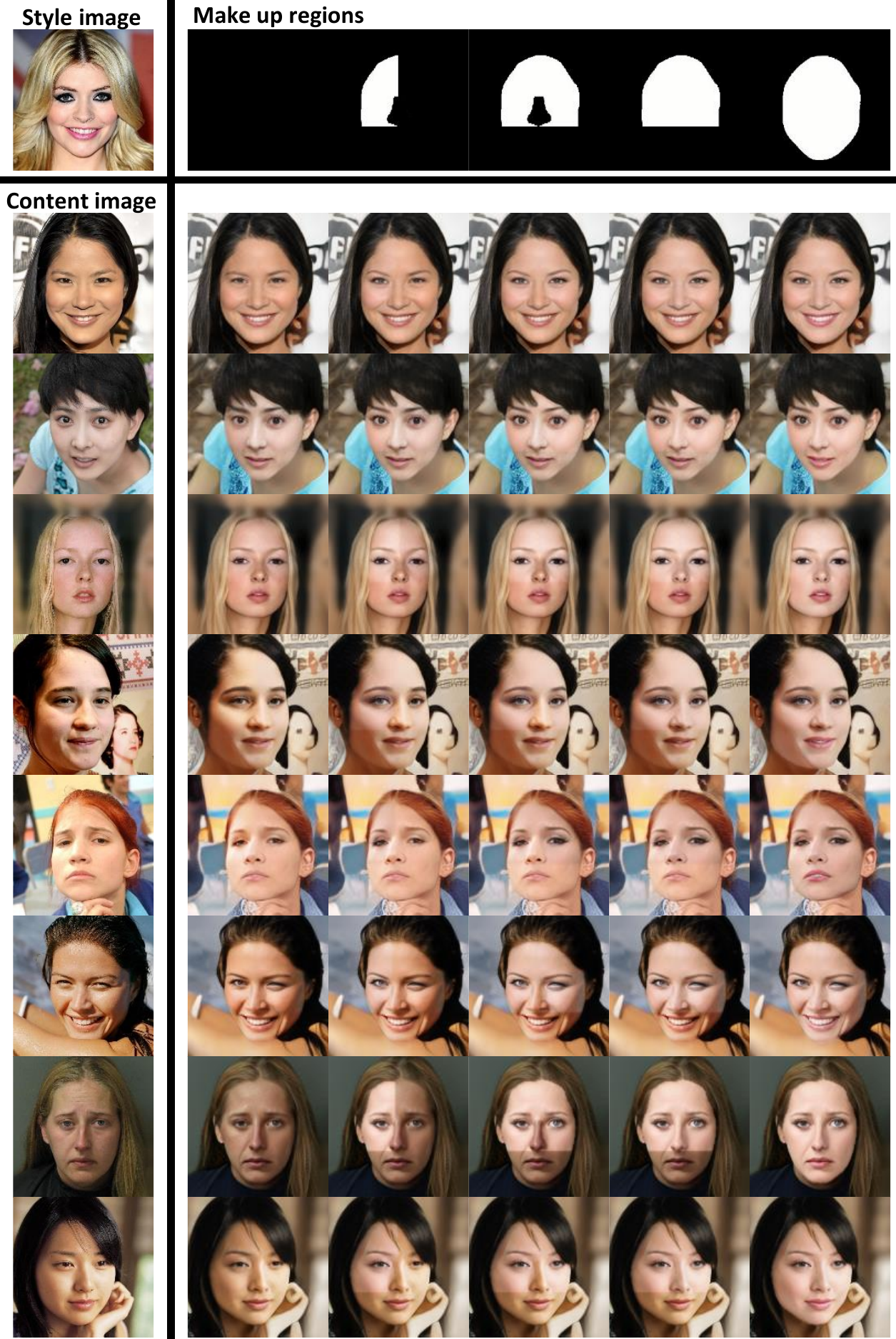}
	\caption{Facial make up step by step using our RFT model.}
	\label{makeup}
\end{figure*}

\end{document}